\documentclass{article}

\usepackage{PRIMEarxiv}

\usepackage[utf8]{inputenc} % allow utf-8 input
\usepackage[T1]{fontenc}    % use 8-bit T1 fonts
\usepackage{hyperref}       % hyperlinks
\usepackage{url}            % simple URL typesetting
\usepackage{booktabs}       % professional-quality tables
\usepackage{amsfonts}       % blackboard math symbols
\usepackage{nicefrac}       % compact symbols for 1/2, etc.
\usepackage{microtype}      % microtypography
\usepackage{lipsum}
\usepackage{fancyhdr}       % header
\usepackage{graphicx}       % graphics
\graphicspath{{media/}}     % organize your images and other figures under media/ folder

\newcommand{\rmnum}[1]{\romannumeral #1}
\newcommand{\tabincell}[2]{\begin{tabular}{@{}#1@{}}#2\end{tabular}}
\makeatother
\usepackage{newfloat}
\usepackage{listings}
\usepackage{amsmath}
\usepackage{multirow}
\usepackage[ruled,linesnumbered]{algorithm2e}
\usepackage{pifont}

\title{Data and Knowledge Co-driving for Cancer Subtype Classification on Multi-Scale Histopathological Slides
\thanks{\textit{\underline{Citation}}: 
\textbf{Yu B, Chen H, Zhang Y, et al. Data and knowledge co-driving for cancer subtype classification on multi-scale histopathological slides[J]. Knowledge-Based Systems, 2023, 260: 110168.}} 
}

\author{
  Bo Yu \\
  School of Artificial Intelligence \\
  Jilin University \\
  Changchun, China\\
   \And
  Hechang Chen \\
  School of Artificial Intelligence \\
  Jilin University \\
  Changchun, China\\
  \texttt{chenhc@jlu.edu.cn} \\
  Corresponding Author \\
   \And
  Yunke Zhang \\
  School of Artificial Intelligence \\
  Jilin University \\
  Changchun, China\\
   \And
  Lele Cong \\
  Department of Neurology \\
  China-Japan Union Hospital of Jilin University \\
  Changchun, China\\
  \texttt{congll18@mails.jlu.edu.cn} \\
  Corresponding Author \\
   \And
  Shuchao Pang \\
  School of Cyber Science and Engineering \\
  Nanjing University of Science and Technology \\
  Nanjing, China\\
   \And
  Hongren Zhou \\
  School of Artificial Intelligence \\
  Jilin University \\
  Changchun, China\\
   \And
  Ziye Wang \\
  Department of Computer Science and Technology \\
  Tongji University \\
  Changchun, China\\
   \And
  Xianling Cong \\
  Tissue Bank \\
  China-Japan Union Hospital of Jilin University \\
  Changchun, China\\
}

\begin{document}
\maketitle

\begin{abstract}
Artificial intelligence-enabled histopathological data analysis has become a valuable assistant to the pathologist. 
However, existing models lack representation and inference abilities compared with those of pathologists, especially in cancer subtype diagnosis, which is unconvincing in clinical practice.
For instance, pathologists typically observe the lesions of a slide from global to local, and then can give a diagnosis based on their knowledge and experience. 
In this paper, we propose a Data and Knowledge Co-driving (D\&K) model to replicate the process of cancer subtype classification on a histopathological slide like a pathologist.
Specifically, in the data-driven module, the bagging mechanism in ensemble learning is leveraged to integrate the histological features from various bags extracted by the embedding representation unit. 
Furthermore, a knowledge-driven module is established based on the Gestalt principle in psychology to build the three-dimensional (3D) expert knowledge space and map histological features into this space for metric. 
Then, the diagnosis can be made according to the Euclidean distance between them.
Extensive experimental results on both public and in-house datasets demonstrate that the D\&K model has a high performance and credible results compared with the state-of-the-art methods for diagnosing histopathological subtypes.
Code: \url{https://github.com/Dennis-YB/Data-and-Knowledge-Co-driving-for-Cancer-Subtypes-Classification}
\end{abstract}

% keywords can be removed

\section{Introduction}
Cancer is among the leading causes of death worldwide, and various cancer subtypes due to gene mutations make it difficult to treat with a universal plan \cite{cancer1}.
Today, innovative research, such as precision medicine, is uncovering proper treatment for each patient based on cancer subtypes to improve the survival rate. 
The diagnosis of cancer subtypes, such as basal cell carcinoma (BCC), squamous cell carcinoma (SCC), and Bowen's disease (BD) in skin cancer (SC), relies heavily on the visual interpretation of histopathological slides by pathologists who use their experience in clinical recognition to render a diagnosis \cite{cancer2}. 
The problem of inconsistency is usually encountered because the diagnosis given by different doctors may not be the same. 
Deep learning (DL) has recently gained much attention as an effective prediction method for classification tasks. 
It is a branch of artificial intelligence (AI) that uses convolutional neural networks (CNNs) \cite{cancer4}, recurrent neural networks (RNNs) \cite{cancer5}, etc., to simulate the process of human decision-making from different aspects. 
In particular, many recent breakthroughs in biomedicine have been in the realm of CNN-driven diagnostics \cite{cancer6,cancer7}.
%, which can assist pathologist to make an objective and accurate decision. 
Therefore, exploring CNN-based methods for histopathological slides is of great significance for the development of cancer subtype classification. 

%So, it is a considerable challenge for fairness and availability because the diagnosis from different doctors may not coincide. 
%Deep learning (DL), a branch of artificial intelligence (AI) that uses artificial neural networks (ANN) \cite{cancer3}, convolutional neural networks (CNN) \cite{cancer4}, recurrent neural networks (RNN) \cite{cancer5}, etc., to simulate aspects of human decision-making, has gained much attention in recent years. 
%Many recent breakthroughs about biomedical have been in the realm of CNN-driven diagnostics \cite{cancer6,cancer7}, which can assist researchers in making an objective and accurate decision. 
%Therefore, adopting the CNN for subtypes classification is significant on the histopathological slide.

% With the development of medical techniques, precision medicine is a newer way to find the proper treatment for each patient based on cancer subtypes to improve the survival rate. 
% The diagnosis of cancer subtypes, particularly solid tumors, relies heavily on the visual interpretation of histopathological slides by pathologists who use their experience in clinical recognition to render a diagnosis \cite{cancer2}. 
% It may happen that the diagnosis results from different doctors do not coincide, which is a considerable challenge for the fairness and availability of diagnosis. 

\begin{figure}[tb]
\centering\includegraphics[width=0.4\textwidth]{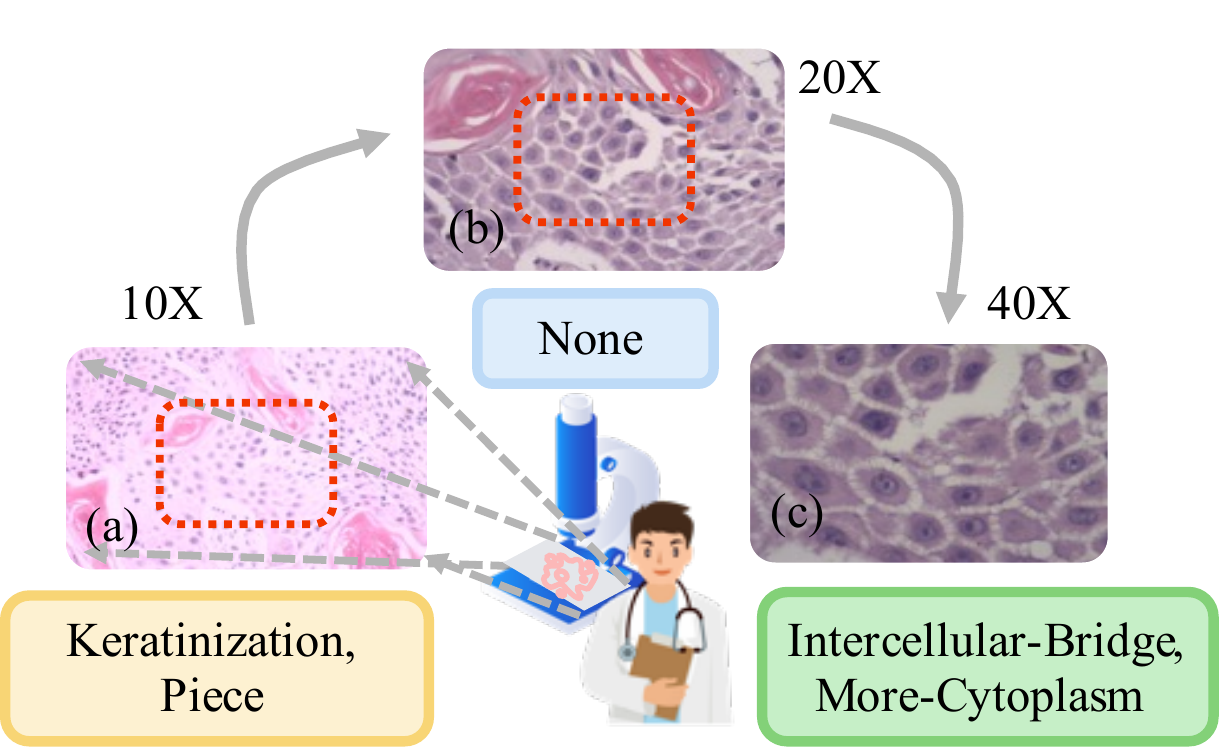}
\caption{The diagnosis process of cancer subtypes by doctors using histopathological images. (a) Cells appear by the piece and begin to keratinize on 10X; (b) There are no other changes on 20X; (c) There are more cytoplasm and intercellular bridges on 40X. Therefore, the subtype is SCC.} 
\label{introduction}
\end{figure}

Previous studies regarding slide classification with CNNs can be roughly divided into whole-based, patch-based, attention-based, and multi-scale methods. 
1) Whole-based methods regard the diagnosis of slides as a classification task of computer vision \cite{pathology1,whole-baseline}. 
%\cite{pathology1,pathology2}
These methods first resize the histopathological slides and then use neural networks to extract features. 
2) Patch-based methods crop the slides into many patches to overcome the problem of information loss \cite{patch2,patch-baseline}. %\cite{patch2,patch5}
The development of multi-instance learning (MIL) \cite{mil-baseline} allows algorithms to find actual tumor regions in these patches and improves classification performance. %\cite{mil1}
3) Attention-based methods utilize the attention mechanism \cite{attention} to extract meaningful information from feature space or channels and guide the network to focus on the regions that significantly affect classification performance \cite{attention-baseline,c2c-baseline}. %\cite{attention1,attention2}
4) Multi-scale methods incorporate contextual information about histopathological slides using multi-scale inputs or extracting hierarchical features \cite{multi-scale-baseline,ms-da-mil}, and they have become the mainstream algorithm of slide classification. %\cite{multi-scale4,multi-scale5}

Despite the success that CNN-driven methods have achieved in classifying histopathological slides, there remain two challenges in subtype classification: 
\textbf{\emph{Challenge I:}} what features should be extracted by the model?
Most current works use CNNs to extract features from input single-scale or multi-scale images \cite{pathology1, patch2, attention-baseline, multi-scale-baseline}, but these high-dimensional features do not have explicit meanings. 
As shown in Figure \ref{introduction}, the lesion varies on the different scales of histopathological slides, e.g., keratinization in 10X and the intercellular-bridge in 40X. 
% An experienced doctor can accurately diagnose the lesions observed globally and locally.
An experienced doctor can accurately diagnose these lesions by observing the slide in a global to local manner.
% However, existing models cannot hierarchically capture these features since they do not discuss what features should be used for diagnosing at different scales, making the diagnosis less interpretable. 
However, existing models cannot capture and represent those complex hierarchical features, leading to an ambiguous feature representation. 
\textbf{\emph{Challenge II:}} how should the model predict the subtypes?
Most methods borrow techniques from computer vision to classify the subtype, e.g., using the softmax function to activate single-scale features, concatenating, or adding multi-scale features for fusion \cite{cvpr2019,ms-da-mil}.
Due to these methods having little understanding of medical knowledge, the diagnosis process is unconvincing.
For example, BD and SCC have similar histological features on slides. 
However, lesions of BD will only appear in the epidermis, whereas SCC may appear in the epidermis or dermis. 
% The doctor can judge based on their knowledge, but the model does not have this medical knowledge, which makes the classification performance of the model insufficient. 
The doctor can infer based on standards, but existing AI models for histopathology diagnosis lack this medical knowledge. Therefore, it creates an unintelligible inference in classifying subtypes. 

In view of this, we propose a data and knowledge co-driving (D\&K) model, which incorporates data-driven and knowledge-driven modules to address these challenges.
% Specifically, for \textbf{\emph{challenge \rmnum{1}}}, the embedding representation stage is adopted to extract bag-level histological features at 10X, 20X, and 40X scales by MIL. 
Specifically, for \textbf{\emph{challenge I}}, we designed an embedding representation unit to extract the histological features of each bag. Then, the model can obtain the histological features of a slide at multiple scales through the bagging ensemble unit. Therefore, the model learns which histological features to use and leverages these explicit representations for diagnosis. Since these two units are trained on data and labels, we call this part the data-driven module.
% Then, the bagging ensemble part converts continuous values into discrete and draws on the voting mechanism in ensemble learning to obtain the final slide-level histological features, which makes the diagnosis results interpretable. 
For \textbf{\emph{challenge II}}, the expert knowledge space is built into the model by the Gestalt principle, a cognitive psychology method that understands different content from the macro to the micro. 
Then, the diagnosis can be made according to the distance of histological prediction results in 3D space.
The model visualizes an intelligible inference for diagnosis by combining expert knowledge, which provides a novel way to improve the performance of classifying subtypes. Therefore, we call this part the knowledge-driven module. In summary, the contributions of this paper are as follows: 
\begin{itemize}
\item 
A novel model called D\&K is proposed for cancer subtype classification based on histopathological slides.
It is a high-performance model incorporating connectionism and symbolism, which is co-driven by the data-driven and knowledge-driven modules.
%We propose a novel D\&K model for high-performance histopathological diagnosis of cancer subtypes, which is the first to explain the diagnosis results by combining connectionism (data) and symbolism (knowledge) in the AI. 

\item 
In the data-driven module, we first design an embedding representation unit to extract histological features from bag-level data. 
Then, a bagging ensemble unit is designed to obtain slide-level histological features, improving the representational ability of the model.
%In the data-driven module, the embedding representation unit is used to extract histological features from the bag-level data. 
%The bagging ensemble unit discretizes features and uses a voting mechanism to obtain slide-level histological representations. 
%In the data-driven module, the embedding representation unit is used to extract histological features from the bag-level data. 
%Then, the bagging ensemble unit discretizes features and then uses a voting mechanism to obtain slide-level histological representations for interpretability. 

\item 
In the knowledge-driven module, diagnostic knowledge of cancer subtypes is embedded into a 3D space by the proximity principle of Gestalt. 
Afterward, a diagnosis can be made by calculating the distance with the similarity principle of Gestalt, thus modeling a high-performance and intelligible inference process.

\item 
Extensive experiments on public and in-house datasets compared with the state-of-the-art methods demonstrate the effectiveness of the D\&K. 
In addition, the visualization process provides more in-depth analyses to illustrate its credible diagnosis results.
\end{itemize}

The remaining paper is organized as follows. 
Section 2 reviews previous work on histopathological application by AI. 
Section 3 and 4 present our solution for classifying subtypes with the data-driven and knowledge-driven modules. 
Experiments and detailed analysis are presented in Section 5, and Section 6 provides the conclusion of the paper.

\section{Related Work}
This section briefly reviews some prior works closely related to our model from four aspects: whole-based, patch-based, attention-based, and multi-scale methods. 
% As many studies utilize CNN for histopathological analysis, we introduced four baseline groups into our work: whole-based, patch-based, attention-based, and multi-scale methods. 

\textbf{Whole-based Methods.}
The straightforward idea is to send the slide directly to the neural network for training \cite{pathology2,wsi2}. 
For instance, Hou et al. developed and used GoogLeNet as a feature extractor to classify breast cancer \cite{googlenet,whole-baseline}. Hekler et al. applied ResNet to implement automatic classification for melanoma histopathological images \cite{resnet,wsi1}. 
However, they usually resize or crop the slides to a standard size appropriate for the model due to the high resolution, which loses the semantic and contextual information.
Thus, the extracted features lack semantic and contextual information, caused by image compression or damage.

\textbf{Patch-based Methods.}
Currently, patch-based methods are more popular, and the image can be divided into many small patches for convenient analysis \cite{patch1,patch3,patch5}. 
For example, Graham et al. and Xu et al. considered that patches contain much detailed information, which is suitable for CNNs to diagnose cancer \cite{patch-baseline,patch4}. 
Among them, MIL can convert a high-resolution image into many patches with bags \cite{mil-baseline,mil4}. 
As a type of supervised learning, MIL receives a set of labeled bags. Each bag contains many instances that do not need to be labeled. For example, the bag will be marked as negative in binary classification if all the instances contained are negative. In contrast, if there is at least one positive instance in a bag, the bag is marked as positive. A mainstream approach is to use the VGG16 network \cite{vgg} as a feature extractor for MIL. During the training, the model tries to find a way to recognize positive instances in a bag and exclude the interference of other negative samples.
However, many negative samples make the extracted features have considerable noise, and it is unclear how the model can distinguish which features are beneficial for classification. 

\textbf{Attention-based Methods.}
The attention mechanism was first used in natural language processing (NLP), enabling a neural network to focus on a subset of its inputs (or features) and select specific inputs \cite{attention-nlp}. 
Its successful application in image classification makes it commonly used in histopathological data \cite{attention}; for example, many researchers use attention mechanisms to guide the CNN to focus on some essential features of histopathological sections, and the features with high attention scores have significant implications for diagnosis \cite{attention3,attention4,attention-baseline}. 
Although the attention mechanism dramatically improves the classification accuracy on these tasks, the features it focuses on may not be essential for distinguishing disease subtypes. 

\begin{figure*}[ht]
\centering\includegraphics[width=0.95\textwidth]{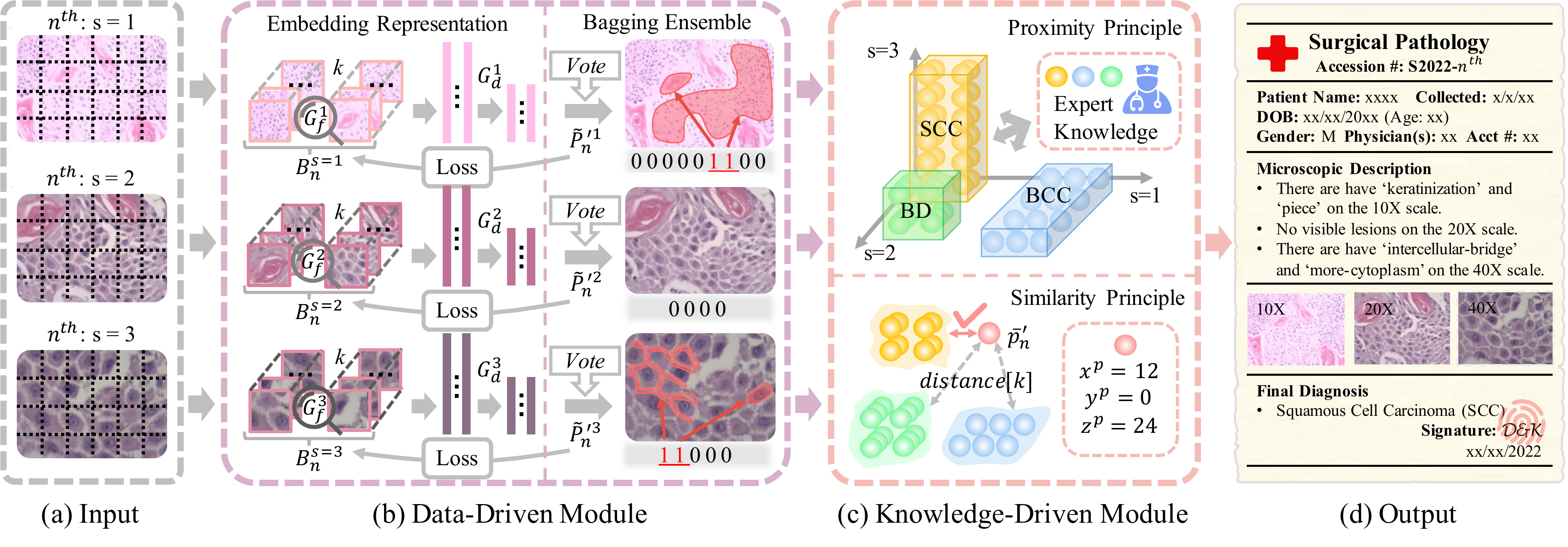}
\caption{An overview of D\&K architecture. (a) Input: Multi-scale images of a slide increase magnification from top to bottom. (b) Data-Driven Module: The histological features of each bag are trained by embedding representation units. Then, the multi-scale histological features of the slide are generated using the bagging ensemble unit. (c) Knowledge-Driven Module: The red dot represents the prediction. Yellow, blue, and green points represent expert knowledge for diagnosis. (d) Output: D\&K predicts the subtype according to the distance and displays the histological features at each scale on the report.} 
\label{architecture}
\end{figure*}

\textbf{Multi-scale Methods.}
Multi-scale methods can improve classification performance by containing contextual information \cite{multi-scale1,multi-scale2}.  %\cite{multi-scale1,multi-scale2,multi-scale3,multi-scale6}. 
For example, some methods exploit the average \cite{COMPSAC2019} or concatenation \cite{cvpr2019} for fusion. 
Hashimoto et al. applied the attention and domain adaptation mechanism for multiple scales \cite{ms-da-mil}. 
Le Trinh et al. proposed the multi-feature method with a binary pattern encoder \cite{multi-scale-baseline}. 
Although these multi-scale methods are commonly used in computer vision, they cannot effectively classify cancer subtypes in histopathological images with complex information. 

% The core idea of our work is inspired to solve \textbf{\emph{challenge \rmnum{1}}} and \textbf{\emph{challenge \rmnum{2}}} in the face of classifying subtypes on histopathological slides. 
% Thus, we propose the D\&K model with two core ideas for solving these challenges. 
Whole-based, patch-based, and attention-based methods all explore how to extract more valuable features, but they do not explicitly state which features are necessary, i.e., \textbf{\emph{challenge I}}.
The multi-scale approach uses multiple pieces of information to predict subtypes, but its diagnostic process is unintelligible due to a lack of expert knowledge, i.e., \textbf{\emph{challenge II}}. The core idea of D\&K is inspired by these challenges for classifying subtypes on histopathological slides. 
Thus, we use the data-driven module to analyze the histological features from global to local, and the knowledge-driven module can judge the subtypes by calculating the distance using the Gestalt principle. 

\section{Problem Statement}
In this section, we use the mathematical formula to define the variables that need to be used in D\&K. 
We let $\{\mathbb{X}_n^s,\mathbb{Y}_n^s,\mathbb{Y}_n,B_n^s,I_{nb}^s\}_{n=1}^{N}$ be a training dataset, where $s$ indicates different scales, e.g., s = 1, 2 or 3 correspond to 10X, 20X or 40X. $N$ is the size of the training dataset, $n\in N$. $\mathbb{X}_n^s$ represents the input image on the $s$ scale of $n^{th}$ slide, and $\mathbb{Y}_n^s$ and $\mathbb{Y}_n$ represent two kinds of labels in the proposed model: histological feature labels on the $s$ scale and subtype labels of the $n^{th}$ slide.
The set of bags from $\mathbb{X}_n^s$ are denoted by $B_n^s$,
% and each bag $b\in B_n^s$ has a set of patches, we denote them as $I_{nb}^s$. 
and we use $I_{nb}^s$ to describe a set of patches in each bag $b\in B_n^s$.
We let $\{\mathbb{X}_n^{'s},\mathbb{Y}_n^{'s},\mathbb{Y}_n^{'},B_n^{'s},I_{nb}^{'s}\}_{n=1}^{N'}$ be a testing dataset, and the meaning of the variables is the same as that of the training dataset.

Given the prediction and the label of histological features $\hat{P}_{nb}^s$ and $\mathbb{Y}_n^s$, we take advantage of three neural networks $G_{f}^{s},G_{d}^{s}$ and $G_{c}^{s}$ to minimize their loss for training. During inference, we obtain the histological features $\widetilde{P}_{n}^{'s}$ from a test slide. 
Then, we use the function $D$ to calculate the distance between expert knowledge and $\widetilde{P}_{n}^{'s}$ to diagnose subtype $\overline{P}_{n}^{'}$, and we can evaluate the performance of the model by comparing the similarity between $\overline{P}_{n}^{'}$ and $\mathbb{Y}_n$.

\section{Methodology}
In this section, we elaborate on D\&K with four subsections in terms of the overall structure, two core modules, and loss function. 

\subsection{The D\&K Architecture}
We introduce the components of D\&K according to Figure \ref{architecture}, and it consists of four parts. 
\emph{Part \rmnum{1}:} We use patch images from three scales of $s = 1$, $s = 2$ and $s = 3$ as input (Figure \ref{architecture} (a)). 
\emph{Part \rmnum{2}:} We use the embedding and bagging mechanism to imitate the pathologist diagnosis process to extract histological features on different scales, e.g., Keratinization and Piece (Figure \ref{architecture} (b)). 
\emph{Part \rmnum{3}:} The model maps histological features into the 3D expert knowledge space and then calculates the Euclidean distance between them for classification (Figure \ref{architecture} (c)). 
\emph{Part \rmnum{4}:} The model generates multi-scale histological features of the slide and predicts the subtypes based on the similarity with knowledge (Figure \ref{architecture} (d)). 
Next, we will describe in detail \emph{Part \rmnum{2}.} and \emph{Part \rmnum{3}.}, which are the two essential components of D\&K.

\begin{algorithm}[tb]
  %\KwData{Skin cancer histopathological image $\{(\mathbb{X}_n^s,\mathbb{Y}_n^s,\mathbb{Y}_n)\}_{n=1}^{N}$}
  \KwIn{$I_{nb}^s$}
  \KwOut{$\widetilde{P}_{n}^{'s}$}
  %\KwResult{Training model parameters}
  Initialization : $G_f^s$, $G_d^s$, $G_c^s$\;
  \% Embedding Representation (Training)\\
  \For{s = 1 to 3}
    {\% Training for E epoch\\
    \For{e = 1 to E}
        {\For{n = 1 to N}
            {
            $H_{nb}^s = G_f^s(I_{nb}^s;\theta_f^s)$\;
            $L_{nb}^s = G_d^s(H_{nb}^s;\theta_d^s)$\;
            $M_{nb}^s = \frac{1}{\mid{L_{nb}^s\mid}}\sum(L_{nb}^s)$\;
            $P_{nb}^s = G_c^s(M_{nb}^s;\theta_c^s)$\;
            $\hat{P}_{nb}^s = Sigmoid(P_{nb}^s)$\;
            $loss = BCELoss(\hat{P}_{nb}^s, \mathbb{Y}_n^s)$\;
            $Backpropagating\ to\ train\ {G}_{f}^{s}, {G}_{d}^{s}, {G}_{c}^{s}$\;
            }
        }
    }
  \% Bagging Ensemble (Testing)\\
  \For{s = 1 to 3}
    {\For{n = 1 to $N^{'}$}
        {
        \% $\theta_f^s$, $\theta_d^s$ and $\theta_c^s$ is fixed\\
        $\hat{P}_{nb}^{'s} = G_c^s(G_d^s(G_f^s(I_{nb}^{'s};\theta_f^s);\theta_d^s);\theta_c^s)$\;
        $\widetilde{P}_{n}^{'s} = Vote((\hat{P}_{nb}^{'s})>v)$\;
        }
    }
  \caption{Data-Driven Module}
  \label{train_algorithm}
\end{algorithm}

\subsection{Data-Driven Module}
Histology is a morphological science that studies the microstructure and functional relationship of an organism. For a slide, the attention of pathologists will change at different scales, which have unique information for analysis.
In general, the scale of $s = 1$ is used to observe the morphological properties of tumor tissues, and cell-level changes can be seen at $s = 2$, while there are nuclear-related traits at $s = 3$. 
Therefore, to address \textbf{\emph{challenge I}}, we propose a data-driven module with the embedding representation and the bagging ensemble to extract multi-scale histological features. Specifically, we represent these features by different binary combinations on three scales. For example, in renal cell cancer (RCC) and skin cancer (SC) datasets used in this paper, their histological features and sequence are as follows: \emph{\textbf{RCC}, s = 1 (ID: 1-6): Nest (Ne), Acinus (Ac), Papillary (Pa), Tubular (Tu), Wall-Thickness (WT), Trabeculae (Tr); s = 2 (ID: 1-6): Thin-Reticulate (TR), Clear-Cell (CC), Foamy (Fo), Psammoma (Ps), Flocculence (Fl), Clear-Boundary (CB); s = 3 (ID: 1-6): Homogenous-Chromatin (HC), Round-Nucleus (RN), Small-Cell (SCe), Double-Nucleus (DN), Irregular-Dark-Nucleus (IDN), Air-Gap-Nucleus (AGN)}; \emph{\textbf{SC}, s = 1 (ID: 1-9): Stripe (St), Ribbon (Ri), Cribriform (Cr), Pigment (Pig), Gap (Ga), Keratinization (Ke), Piece (Pie), Hypertrophy (Hy), Epidermis (Ep); s = 2 (ID: 1-4): Interstitial (In), Fence (Fe), Implicate-Vessel (IV), Implicate-Adnexa (IA); s = 3 (ID: 1-5): Intercellular-Bridge (IB), More-Cytoplasm (MC), Nuclear-Vacuolation (NV), Cytoplasm-Vacuolation (CV), Alien-Cell (AC)}. We regard the modeling as a multi-label classification task consisting of these two steps, and Algorithm \ref{train_algorithm} describes the modeling process of the data-driven module. Due to $E$ being a constant, the time and space complexities are $O(n)$; please see details in Appendix B.

\textbf{Embedding Representation.} 
First, we use VGG16 with MIL as the extractor to obtain high-dimensional embeddings, which can reflect the morphological changes of slides at each scale and guide the neural network to learn and identify important histological features for classification. 
After the patches $I_{nb}^s$ are processed by the extractor, the output $H_{nb}^{s}$ is $k\times Q$-dimensional vectors, where $k$ is the number of patches in a bag, and $Q$ is the number of features for a patch. 
Thus, we can define this process as follows:
\begin{equation}
    H_{nb}^{s} = G_f^s(I_{nb}^s;\theta_f^s),
    \label{h}
\end{equation}
where $G_f^s$ represents the extractor, and $\theta_f^s$ is the parameter of $G_f^s$ to be optimized. 
Since the dimension of $H_{nb}^s$ is large, a fully connected network of $G_d^s$ is used to reduce the dimension of $H_{nb}^s$. 
$L_{nb}^s$ is obtained with $k\times 512$ dimensions after being processed by $G_d^s$. 
It can be defined as:
\begin{equation}
    L_{nb}^s = G_d^s(H_{nb}^s;\theta_d^s).
    \label{l}
\end{equation}
Similarly, $\theta_d^s$ is the parameter of $G_d^s$ to be optimized. Then, we use the averaged method to fuse patches, and this method can be expressed as: 
\begin{equation}
    M_{nb}^s = \frac{1}{\mid{
    L_{nb}^s\mid}}\sum(L_{nb}^s).
    \label{mean}
\end{equation}
In Equation \eqref{mean}, $M_{nb}^s$ is a $1\times512$-dimensional vector, and we send $M_{nb}^s$ to the classification layer to obtain the representation of histological features at each scale. 
It can be expressed by the following:
\begin{equation}
    P_{nb}^s = G_c^s(M_{nb}^s;\theta_c^s),
\end{equation}
where $G_c^s$ is a fully connected neural network and $\theta_c^s$ is the parameters of $G_c^s$. 
Next, we feed it into the $Sigmoid$ activation function to obtain the multi-label prediction of $\hat{P}_{nb}^s$. 
Namely, they represent the probability of the histological features under the $s$ scale, and we send $\hat{P}_{nb}^s$ and $\mathbb{Y}_n^s$ into the loss function of $\mathcal{L}$ as follows:
\begin{equation}
    \theta_f^s, \theta_d^s, \theta_c^s \leftarrow argmin\sum_{n=1}^{N}\sum_{b\in B_n^s}\mathcal{L}(\hat{P}_{nb}^s, \mathbb{Y}_n^s).
    \label{bp}
\end{equation}
In Equation \eqref{bp}, the model uses gradient descent to minimize the loss value for optimizing the parameters.

\begin{table}[t]
\centering
\footnotesize
\caption{An example of RCC expert knowledge on $s = 1$. KIRC: kidney renal clear cell carcinoma, KIRP: kidney renal papillary cell carcinoma, KICH: kidney chromophobe.}
\setlength{\tabcolsep}{3.5mm}{
\begin{tabular}{ccccccc}
\toprule
\rotatebox{0}{Subtypes} & \rotatebox{0}{Ne} & \rotatebox{0}{Ac} & \rotatebox{0}{Pa} & \rotatebox{0}{Tu} & \rotatebox{0}{WT}& \rotatebox{0}{Tr}\\
\midrule
\midrule
\multirow{3}*{\rotatebox{0}{KIRC}} & 1& 1& 0& 0& 0& 0 \\
		 & 1& 0& 0& 0& 0& 0 \\
		 & 0& 1& 0& 0& 0& 0 \\
\midrule
\multirow{3}*{\rotatebox{0}{KIRP}} & 0& 0& 1& 1& 0& 0 \\
		 & 0& 0& 1& 0& 0& 0 \\
		 & 0& 0& 0& 1& 0& 0 \\
\midrule
\multirow{3}*{\rotatebox{0}{KICH}} & 0& 0& 0& 0& 1& 1 \\
		 & 0& 0& 0& 0& 1& 0 \\
		 & 0& 0& 0& 0& 0& 1 \\
\bottomrule
\end{tabular}
}
\label{table3}
\end{table}

\textbf{Bagging Ensemble.} 
As mentioned above, we obtain many bag-level histological features through the trained embedding representation unit. These bag-level features should be aggregated for one slide in the test stage. 
Therefore, each bag needs to obtain discrete values of histological features to indicate whether the feature is present in the image, which will be further combined into a slide representation. 
We define $v$ as the label threshold, and it is a hyperparameter that can change the representation of the bag from continuous to discrete. 
For example, if an element in $\hat{P}_{nb}^{'s}$ is larger than $v$, the value is 1; otherwise, it is 0. 
The default value of $v$ is 0.5, and it will fluctuate due to the number of slides and the occurrence frequency of histological features in the dataset, and it will be analyzed in detail in the experimental section. 
Next, the model needs to integrate the results of bags and calculate the unique value of all $\hat{P}_{nb}^{'s}$, which can obtain a representation for the slide on this scale.
\begin{equation}
    \widetilde{P}_{n}^{'s} = \sum_{b=1}^{B_n^{'s}}((\hat{P}_{nb}^{'s})>v)>\frac{B_n^{'s}}{2}.
    \label{sample_pred}
\end{equation}
We refer to the bagging algorithm in ensemble learning \cite{ensemble} to vote on multiple bags for one slide. In Equation \eqref{sample_pred}, if more than half of the bags think that the histological feature appears at the $s$ scale, the corresponding histological feature value is set to 1; otherwise, it is set to 0. 
These predicted histological features provide a basis for subsequent diagnosis, which allows D\&K to have a powerful ability to represent the features.

\begin{algorithm}[tb]
  %\KwData{Skin cancer histopathological test image $\{(\mathbb{X}_n^{'s},\mathbb{Y}_n^{'s},\mathbb{Y}_n^{'})\}_{n=1}^{N'}$}
  \KwIn{$\widetilde{P}_{n}^{'s}$}
  \KwOut{$\overline{P}_{m}^{'}$}
  %\KwResult{Subtype prediction results}
  Reload : $G_f^s$, $G_d^s$, $G_c^s$\;
  \% Proximity Principle, EK means expert knowledge\\
  \For{n = 1 to $N^{'}$}
      {
        ${x}^{p}, {y}^{p}, {z}^{p} = Decimal(\widetilde{P}_{n}^{'s=1}, \widetilde{P}_{n}^{'s=2}, \widetilde{P}_{n}^{'s=3})$\;
        \For{j = 1 to 3}
        {
            ${x}^{l}, {y}^{l}, {z}^{l} = Decimal(EK[j][s=1], EK[j][s=2], EK[j][s=3])$\;
            \% Similarity Principle\;
            ${P}_{n}^{'j}.append(\Vert[{x}^{p}, {y}^{p}, {z}^{p}], [{x}^{l}, {y}^{l}, {z}^{l}]\Vert_2)$\;
            ${P}_{n}^{'}.append(Min({P}_{n}^{'j}))$\;
        }
        $\overline{P}_{n}^{'} = IndexMin({P}_{n}^{'})$\;
      }
  \caption{Knowledge-Driven Module}
  \label{prediction_algorithm}
\end{algorithm}

\subsection{Knowledge-Driven Module}
Knowledge is a set of reference lists mentioned in the data-driven module for describing various subtypes, and each list consists of different histological features represented by binary codes. 
As shown in Table \ref{table3}, \emph{Ne} and \emph{Ac} are lesions of clear cell carcinoma on the '$s = 1$' scale, while \emph{Pa} and \emph{Tu} are lesions of papillary cell carcinoma.
They can be represented as (1, 1, 0, 0, 0, 0) and (0, 0, 1, 1, 0, 0), respectively. Therefore, the expert knowledge in this paper can be defined as a binary combination of histological features, and each subtype has independent histological features corresponding to it.
All combinations of knowledge are detailed in Section A of the Appendix.
The doctor can make an accurate diagnosis based on this knowledge, and it is critical to give this knowledge to the model. 
Today, we can easily access and leverage this diagnostic knowledge from doctors and the literature, and they are usually composed of histological features from three scales. Therefore, an ideal way is to map these prediction features into the 3D expert knowledge space for interaction, in which each dimension represents a scale. This method can generate metric spaces for distinguishing subtypes according to the three scale features and fuse data-based results into the expert knowledge space with the whole model. Next, how should we incorporate them into the 3D space? 
The Gestalt principle is a fundamental theory in cognitive psychology that points out that a visual image is first recognized as a global entity and then understood as content specific to the local \cite{gestalt}. It coincides with the diagnosis process by pathologists, and two subprinciples within it can support the fusion process in the model. 
Specifically, the proximity subprinciple illustrates that humans tend to divide things by distance during observation. The similarity subprinciple states that the visual system will classify objects with similar features into the same class \cite{subgestalt}. 
Similarly, expert knowledge from the same subtype should be closer in space, and then the model can generate a diagnosis based on the similarity of histological features at three scales. 
Therefore, we use these subprinciples to build the expert knowledge space, embed histological features into space, and then calculate the distance between them to obtain the subtype. 
The knowledge-driven module is divided into two steps for solving \textbf{\emph{challenge II}}, and Algorithm \ref{prediction_algorithm} describes the prediction process. The time and space complexities are $O(n)$; please see details in Appendix B.

\begin{figure}[tb]
\centering\includegraphics[width=0.45\textwidth]{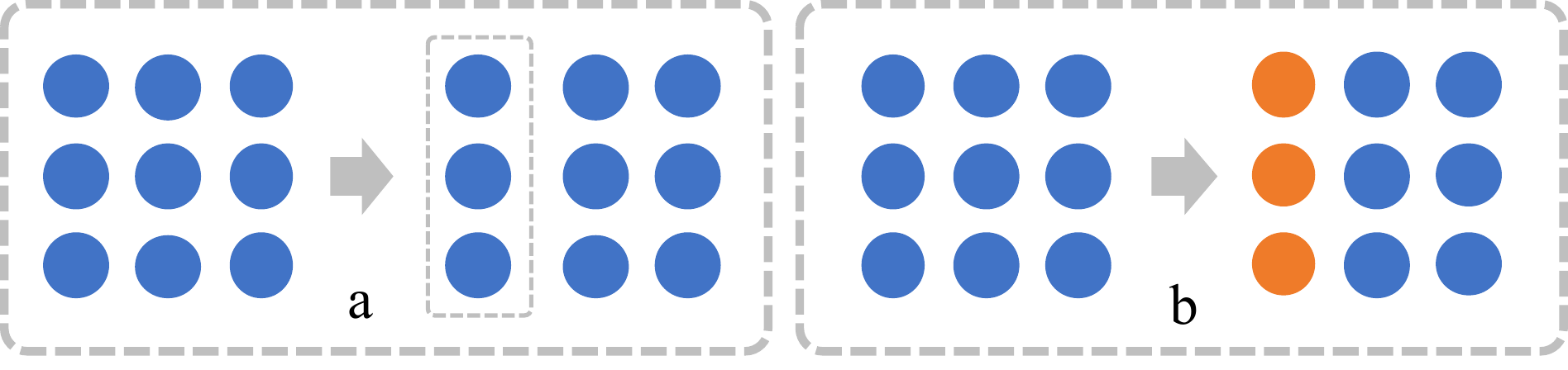}
\caption{Two subprinciples of the Gestalt we used in D\&K. (a) The proximity principle, and (b) the similarity principle.} 
\label{gestalt}
\end{figure}

\textbf{Proximity Principle.} 
The proximity principle emphasizes that we should put information with similar attributes together and separate irrelevant information as much as possible. 
For example, as shown in Figure \ref{gestalt} (a), the three points surrounded by dashed lines are closer than others, so we treat them as a group. 
In other words, if two elements are close, they are more likely to come from the same group, and their relationship is closer than before. 
Similarly, the expert knowledge in the 3D space needs to be grouped according to the subtype, which are easy to distinguish. 
Therefore, we convert the predicted histological features of the binary coded to decimal at each scale, regard them as coordinates in the expert knowledge space, and cluster them together by the same category.
By bagging ensemble unit in the data-driven module, we obtain the histological features $\widetilde{P}_{n}^{'s}$ of the $n^{th}$ slide from $I_{nb}^{'s}$ on the $s$ scale, which can be defined as: 
\begin{equation}
    \widetilde{P}_{n}^{'s} = Vote(G_c^s(G_d^s(G_f^s(I_{nb}^{'s};\theta_f^s);\theta_d^s);\theta_c^s)>v).
    \label{equ_pred}
\end{equation}
Next, they can be regarded as binary coded, and we convert the predicted histological features from binary to decimal on the scale of $s = 1, 2, 3$, and they become the points in 3D expert knowledge space represented by ${x}^{p}$, ${y}^{p}$, ${z}^{p}$ and ${x}^{l}$, ${y}^{l}$, ${z}^{l}$, respectively. 

\textbf{Similarity Principle.} 
According to the similarity principle, things with prominent characteristics (e.g., shape, color, size) should be combined. 
In Figure \ref{gestalt} (b), since the three yellow points have the same color, we treat them as a category. 
Namely, the elements with similar visual characteristics are considered more related. 
In our model, we compare the distribution of the prediction with expert knowledge and find which prediction is more similar to a certain subtype of expert knowledge, and then we tend to predict the sample as this subtype. 
Specifically, the model uses the Euclidean distance to obtain the distance for each subtype, and it can be defined as follows:
\begin{equation}
    D = sqrt(\sum_{pos = x, y, z}({pos}^{p} - {pos}^{l})^{2}).
    \label{dist}
\end{equation}
Equation \eqref{dist} is the process of similarity measurement. For each subtype $j$, we put all distances into ${P}_{n}^{'j}$, and we put the minimal distance from ${P}_{n}^{'j}$ into ${P}_{n}^{'}$, then obtain the index value $\overline{P}_{n}^{'}$ of ${P}_{n}^{'}$, where $\overline{P}_{n}^{'}$ is the subtype for the $n^{th}$ slide. 
Moreover, it can be diagnosed as BD without calculating similarity if the \emph{Ep} feature appears when $s = 1$ in skin cancer. 
In this way, D\&K builds an intelligible inference process and provides a credible result due to the integration of expert knowledge. 

Overall, if we first input the SCC sample, as shown in Figure \ref{introduction}, the data-driven module can then obtain histological features at three scales: (0, 0, 0, 0, 0, 1, 1, 0, 0), (0, 0, 0, 0), and (1, 1, 0, 0, 0). Next, the knowledge-driven module maps these three features in the appendix into the 3D expert knowledge space to make subtype diagnoses based on their similarity. Finally, by calculating the point closer to the SCC class, the model can diagnose the slide as SCC and provide its histology features.

\subsection{Loss Function}
In D\&K, the process of extracting histological features can be regarded as a multi-label classification task. 
Therefore, we use the binary cross-entropy loss function (BCELoss) with the sigmoid function to calculate the loss value, which can be defined as follows: 
\begin{equation}
    \begin{aligned}
        L = -\frac{1}{N}\sum_{n=1}^{N}(\mathbb{Y}_{n}^{s} \cdot log(\hat{P}_{nb}^{s})+ (1-\mathbb{Y}_{n}^{s}) \cdot log(1-\hat{P}_{nb}^{s})),
        \label{loss}
    \end{aligned}
\end{equation}
where $\mathbb{Y}_{n}^{s}$ represents the label of histological features for one sample on the scale $s$, and $\hat{P}_{nb}^{s}$ represents the prediction of the embedding representation unit. 
In addition, we can also use BCE with logits loss function (BCEWithLogitsLoss), which removes the sigmoid function before calculating the loss.

\section{Experiments}
In this section, we first describe the experimental settings and then conduct four groups of experiments to answer the following questions: 
\textbf{Q1}: How does our proposed method perform compared with the state-of-the-art methods? 
\textbf{Q2}: How are histological features represented and integrated with data-driven modules? 
\textbf{Q3}: Is it necessary to combine expert knowledge by the Gestalt principles in the knowledge-driven module? 
\textbf{Q4}: How does D\&K obtain explicit representation and inference ability for improving the performance of subtype classification? 

\subsection{Experimental settings}
\subsubsection{Dataset}
To fairly compare the generality of D\&K for cancer subtype diagnosis, we selected RCC histopathological slides from The Cancer Genome Atlas (TCGA) database \cite{tcga}. 
It is a landmark cancer genomics program that molecularly characterizes over 20,000 primary cancer and matched normal samples spanning 33 cancer types. 
We selected 50 samples from each of the three RCC subtypes at three magnification ratio. The categories were kidney renal clear cell carcinoma (KIRC), kidney renal papillary cell carcinoma (KIRP), and kidney chromophobe (KICH). Therefore, there is a total of 150 RCC samples for a fair comparison, and it is approximately the same number of samples as the in-house dataset we used.

Since the samples in TCGA are more typical, to verify the effect of the algorithm in the real world, we need to distinguish more complex data. 
Based on this, we obtained 145 SC histopathological samples with three scale images from cooperative medical institutions, which contained 60, 55, and 30 samples of BCC, SCC, and BD, respectively. 
The images stained by hematoxylin and eosin (H\&E) were captured using a slide scanner with the ocular lens at 10X and the objective lens in scales of 10X, 20X, and 40X. 
They were certified by ethics committees and allowed to be used in AI-based scientific research. 
The protocol of the study was approved by Jilin University and the China-Japan Union Hospital of Jilin University. 

\begin{figure}[tb]
\centering\includegraphics[width=0.45\textwidth]{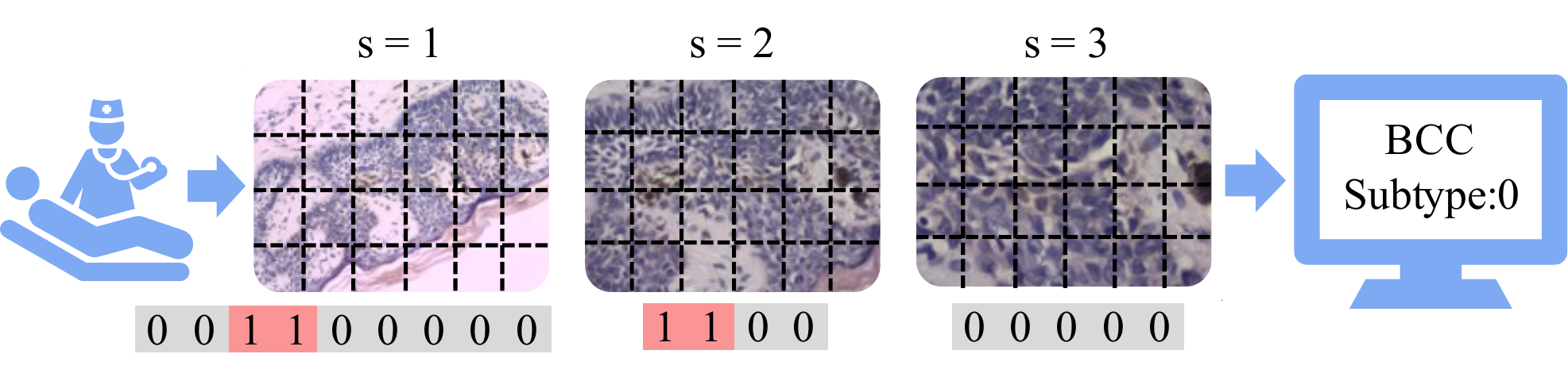}
\caption{Example of inputs and their annotations for one sample. The dividing line represents the patches after cropping. The numbers at the bottom of the image are histological feature labels, and the subtype label is on the right.} 
\label{dataset}
\end{figure}

We take 80\% of the data for training and the rest for testing. 
As shown in Figure \ref{dataset}, there is one subtype label and three histological feature labels for a sample. 
Meanwhile, the histological feature annotations and knowledge for these two datasets come from professional pathologists engaged in the China-Japan Union Hospital of Jilin University. 
Professional pathologists summarized the diagnostic knowledge of subtypes based on relevant literature and clinical experience. This knowledge has been provided in tables, as shown in the Appendix. 
All experiments were performed in accordance with relevant laws and regulations. 

\subsubsection{Baselines and Evaluation}
To comprehensively compare the performance between our method and the state-of-the-art algorithms, we choose seven algorithms from four groups of baselines for verification. 
The public codes of the comparison algorithms are used, and we reproduced the code for unpublished works. 
\begin{itemize}
\item \emph{Whole-based method}: This method uses the whole compressed histopathological image as input to classify positive or negative \cite{whole-baseline}. 

\item \emph{Patch-based method}: The model classifies the patches cropped from the image \cite{patch-baseline}, and the MIL is based on the bag consisting of patches \cite{mil-baseline}. 

% \item \emph{MIL-based method}: By learning multi-instance bags with classification labels \cite{mil-baseline}, a multi-instance classifier is established and applied to predict unknown multi-instance bags. 

\item \emph{Attention-based method}: Attention-based methods \cite{attention-baseline, c2c-baseline} can discover the most critical instances in the group and assign weights to different instances for classification. 

% \item \emph{Domain-based method}: DA-MIL \cite{ms-da-mil} gives a reliable solution to the adaptive problem for different medical institutions. 

\item \emph{Multi-scale method}: MS-DA-MIL \cite{ms-da-mil} leverages multiple inputs for fusion, but MSBP-Net \cite{multi-scale-baseline} uses multiple features for one input, so it has no $s = 1,2,3$. 
\end{itemize}

In addition, we select three strategies from ConvNet \cite{COMPSAC2019}, AWMF-CNN \cite{cvpr2019}, and MS-DA-MIL \cite{ms-da-mil} that combine the multi-scale features from our data-driven module by neural networks to verify the effectiveness when adding expert knowledge by the Gestalt principle. 
We used five evaluation metrics for testing: accuracy(ACC), precision(P), recall(R), specificity(S), and F1-score(F1). %\cite{evaluation}

\subsubsection{Implementation}
%In the training stage, each image is divided into 50 bags, and there are 40 patches in each bag with the size of 224 $\times$ 224. 
The input resolution of each patch is 224 $\times$ 224, and we use VGG16 as the feature extractor with a learning rate of $10^{-4}$ and momentum of 0.9 in the SGD optimizer. 
In the test stage, the numbers of bags and patches are the same as in the training, and the initial value of the label threshold $v$ is set to 0.5. 
All experiments are implemented using PyTorch 1.5.1, Python 3.7, and the Cuda 10.2 framework on an NVIDIA TITAN RTX-24GB GPU and 64GB RAM.

\begin{table*}[tb]
\centering
\caption{The results of different algorithms on RCC and SC datasets. All of these values are from the best experiment results in ten consecutive tests.}
\footnotesize
\setlength{\tabcolsep}{1.5mm}{
%\resizebox{\linewidth}{!}{
\begin{tabular}{cc|ccccc|ccccc}
\toprule
 &  & \multicolumn{5}{c}{RCC} & \multicolumn{5}{c}{SC}\\
\midrule
Method & Scale & ACC & P & R & S & F1 & ACC & P & R & S & F1\\
\midrule
\midrule
\multirow{3}{*}{\tabincell{c}{Whole-based \cite{whole-baseline}\\Hou (2020)}}& s = 1 & 1.0000 & 1.0000 & 1.0000 & 1.0000 & 1.0000 & 0.5862 & 0.5852 & 0.5783 & 0.7815 & 0.5713\\
		 & s = 2 & 1.0000 & 1.0000 & 1.0000 & 1.0000 & 1.0000 & 0.5517 & 0.7348 & 0.5530 & 0.7582 & 0.5492\\
		 & s = 3 & 0.9667 & 0.9697 & 0.9667 & 0.9833 & 0.9666 & 0.4828 & 0.6429 & 0.4344 & 0.7135 & 0.4358\\
\specialrule{0em}{1pt}{1pt}
\midrule
\multirow{3}{*}{\tabincell{c}{Patch-based \cite{patch-baseline}\\Graham et al. (2018)}}& s = 1 & 0.9846 & 0.9847 & 0.9846 & 0.9923 & 0.9846 & 0.5272 & 0.4716 & 0.4487 & 0.7406 & 0.4136\\
		 & s = 2 & 0.9566 & 0.9565 & 0.9566 & 0.9783 & 0.9565 & 0.4173 & 0.3400 & 0.3588 & 0.6914 & 0.3452\\
		 & s = 3 & 0.8195 & 0.8211 & 0.8195 & 0.9098 & 0.8094 & 0.5698 & 0.5597 & 0.5340 & 0.7702 & 0.5401\\
\specialrule{0em}{1pt}{1pt}
\cline{2-12}
\specialrule{0em}{1pt}{1pt}
\multirow{3}{*}{\tabincell{c}{MIL \cite{mil-baseline}\\Couture et al. (2018)}}& s = 1 & 1.0000 & 1.0000 & 1.0000 & 1.0000 & 1.0000 & 0.8290 & 0.8345 & 0.7648 & 0.9067 & 0.7765\\
		 & s = 2 & 1.0000 & 1.0000 & 1.0000 & 1.0000 & 1.0000 & 0.7034 & 0.6797 & 0.6543 & 0.8441 & 0.6573\\
		 & s = 3 & 0.9508 & 0.9572 & 0.9508 & 0.9754 & 0.9506 & 0.7014 & 0.6619 & 0.6507 & 0.8447 & 0.6527\\
\specialrule{0em}{1pt}{1pt}
\midrule
\multirow{3}{*}{\tabincell{c}{Attention-MIL \cite{attention-baseline}\\Ilse et al. (2018)}}& s = 1 & 0.9692 & 0.9718 & 0.9692 & 0.9846 & 0.9691 & 0.7641 & 0.7628 & 0.6895 & 0.8711 & 0.6926\\
		 & s = 2 & 1.0000 & 1.0000 & 1.0000 & 1.0000 & 1.0000 & 0.6145 & 0.4107 & 0.5183 & 0.7870 & 0.4580\\
		 & s = 3 & 0.9417 & 0.9491 & 0.9417 & 0.9709 & 0.9413 & 0.6869 & 0.5322 & 0.5791 & 0.8279 & 0.5264\\
\specialrule{0em}{1pt}{1pt}
\cline{2-12}
\specialrule{0em}{1pt}{1pt}
\multirow{3}{*}{\tabincell{c}{C2C \cite{c2c-baseline}\\Sharma et al. (2021)}}& s = 1 & 1.0000 & 1.0000 & 1.0000 & 1.0000 & 1.0000 & 0.8621 & 0.8983 & 0.8308 & 0.9226 & 0.8518\\
		 & s = 2 & 1.0000 & 1.0000 & 1.0000 & 1.0000 & 1.0000 & 0.7586 & 0.7685 & 0.7727 & 0.8865 & 0.7498\\
		 & s = 3 & 0.9333 & 0.9444 & 0.9333 & 0.9667 & 0.9327 & 0.6897 & 0.4786 & 0.5808 & 0.8319 & 0.5247\\
\midrule
\multirow{4}{*}{\tabincell{c}{MS-DA-MIL \cite{ms-da-mil}\\Hashimoto et al. (2020)}}& s = 1 & 1.0000 & 1.0000 & 1.0000 & 1.0000 & 1.0000 & 0.8386 & 0.8540 & 0.7846 & 0.9118 & 0.7980\\
		 & s = 2 & 1.0000 & 1.0000 & 1.0000 & 1.0000 & 1.0000 & 0.7145 & 0.6514 & 0.6447 & 0.8520 & 0.6394\\
		 & s = 3 & 0.9700 & 0.9721 & 0.9700 & 0.9850 & 0.9699 & 0.7214 & 0.6938 & 0.6839 & 0.8559 & 0.6865\\
         & s = 1,2,3 & 0.9733 & 0.9753 & 0.9733 & 0.9867 & 0.9733 & 0.7372 & 0.6778 & 0.6720 & 0.8649 & 0.6709\\
\specialrule{0em}{1pt}{1pt}
\cline{2-12}
\specialrule{0em}{1pt}{1pt}
\multirow{3}{*}{\tabincell{c}{MSBP-Net \cite{multi-scale-baseline}\\Le Trinh et al. (2021)}}& s = 1 & 1.0000 & 1.0000 & 1.0000 & 1.0000 & 1.0000 & 0.7241 & 0.4853 & 0.6061 & 0.8464 & 0.5367\\
		 & s = 2 & 1.0000 & 1.0000 & 1.0000 & 1.0000 & 1.0000 & 0.7241 & 0.8370 & 0.6339 & 0.8442 & 0.6254\\
		 & s = 3 & 0.8333 & 0.8671 & 0.8333 & 0.9167 & 0.8283 & 0.6552 & 0.4461 & 0.5480 & 0.8072 & 0.4869\\
\midrule
\midrule
\multirow{4}*{D\&K (Ours)} & s = 1 & 1.0000 & 1.0000 & 1.0000 & 1.0000 & 1.0000 & 0.8621 & 0.8500 & 0.8333 & 0.9278 & 0.8390\\
 & s = 2 & 1.0000 & 1.0000 & 1.0000 & 1.0000 & 1.0000 & 0.7241 & 0.4935 & 0.6111 & 0.8497 & 0.5397\\
 & s = 3 & 0.9333 & 0.8333 & 1.0000 & 0.9000 & 0.9327 & 0.7241 & 0.7000 & 0.6894 & 0.8566 & 0.6927\\
 & s = 1,2,3 & \textbf{1.0000} & \textbf{1.0000} & \textbf{1.0000} & \textbf{1.0000} & \textbf{1.0000} & \textbf{0.8966} & \textbf{0.9184} & \textbf{0.8889} & \textbf{0.9434} & \textbf{0.8984}\\
\bottomrule
\end{tabular}
}
\label{table1}
\end{table*}

\subsection{Overall Experimental Results (Q1)}
To verify the effectiveness of our model, we compare the performance of D\&K with state-of-the-art algorithms. 
As shown in Table \ref{table1}, most methods perform well on the RCC subtype data from TCGA. The results of the D\&K and some other models are 1.0000 in multiple metrics of the RCC dataset. This phenomenon shows that the nuance between different subtypes of RCC in TCGA is obvious, and the model can accurately extract relevant features for classification. 
It also shows that TCGA provides data with apparent lesions after screening as a standard public dataset. Therefore, the metrics reached 1.0000 on this dataset due to the small sample size used in this experiment with apparent features. In contrast, the performance of different algorithms on SC data is quite different. It is more complex and challenging because it originates from raw clinical data without screening, which illustrates the importance of analyzing raw data from the real world. 
Next, we mainly compare the performance of the different algorithms on the more complex in-house dataset. 

First, whole-based methods directly send the image to the CNN, which loses more information during preprocessing, and they have a mediocre performance. 
Second, patch-based methods cut the whole image into many patches, and the MIL makes some progress by combining it with many bags. 
Although they retain the detailed information of the image, the results still have no obvious advantages. 
The reason is that they are not designed for histopathological images, which usually makes them ambiguous for what features they should observe. 
In view of this, we address the problem of interpretability with the data-driven module by using MIL and bagging mechanisms to generate histological features. 
Our model can find the lesions of different subtypes, which explicitly express the reason for diagnosis.

Then, attention-based methods can help the network extract essential features and improve classification performance. 
The C2C combined with the attention mechanism achieves significant progress in classification results. 
Finally, MS-DA-MIL and MSBP-Net are all multi-scale methods, and they combine the image by concatenation or feature pyramid.
However, their performance cannot surpass some single-scale algorithms that have been mentioned above. 
The reason is that these methods cannot show which scale features they have used and cannot explain why they combine the features in this way. 
Therefore, we introduce expert knowledge with the Gestalt principle to obtain the best subtype classification results. 
As shown in Table \ref{table1}, our model has gained better performance than baseline methods on both single and multi scales. It should be noted that the D\&K with multi-scale expert knowledge performs better than the single scale, illustrating the importance of combining information from various scales by the Gestalt principle for subtype classification. Specifically, D\&K gained an accuracy of 0.9333 on the third scale from the RCC dataset due to lost expert knowledge for diagnosis from the global perspective. At the same time, the performance of D\&K at the single scale is also more defective than in the multi-scale phase on the SC dataset. All these phenomena show that using the Gestalt principle to fuse multi-scale expert knowledge can achieve better results than the single scale.

In general, D\&K rebuild the diagnosis process as the pathologist by data and knowledge co-driving and surpassed the state-of-the-art methods with 100\% and 89.66\% on these two datasets. 
This demonstrates the effectiveness of D\&K for subtype classification on complex histopathological data from the real world. 

\begin{figure*}[tb]
\centering\includegraphics[width=1\textwidth]{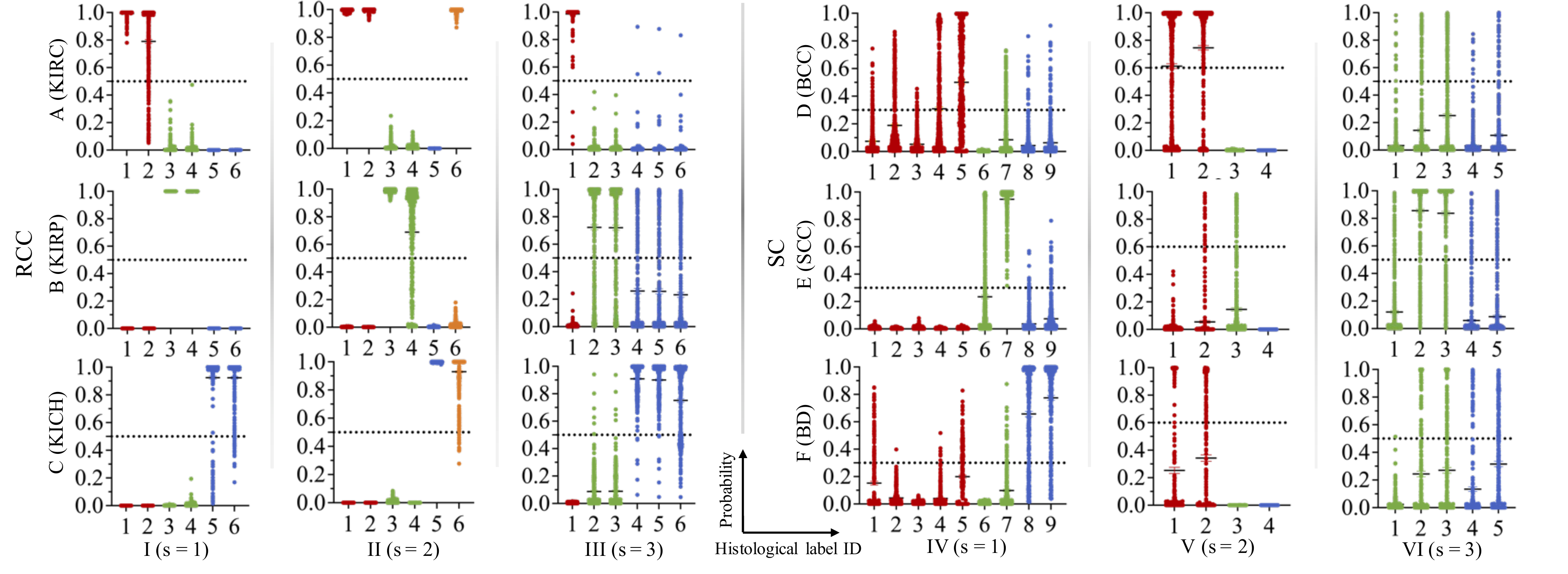}%0.48
\caption{The graph of histological features distribution on the test dataset. The red, green, and blue points correspond to the unique histological features of the subtype, and the orange points are the features shared by more than two subtypes.} 
\label{ensemble}
\end{figure*}

\subsection{Effectiveness of the Data-Driven Module (Q2)}
In this section, we first verify the extraction ability of the embedding representation unit. 
In Figure \ref{ensemble}, the bag-level histological features of different subtypes can be learned well and represented by neural networks, i.e., the features corresponding to the subtypes will have a higher probability. 
For example, in the subgraph of (A-I), the \emph{Ne} (ID: 1) and \emph{Ac} (ID: 2) features have high probability in KIRC, but they rarely appear in KIRP (B-I) and KICH (C-I). 
As seen from Figure \ref{ensemble}, the embedding representation unit can accurately find most of the histological features belonging to subtypes. 
We can obtain discrete histological features of bag-level slides by rationally dividing the label threshold. 
This demonstrates that our embedding representation unit can learn the correct histological feature representation for bags from a large amount of data. 
These histological features can tell us the categories of lesions for bag-level slides, allowing the model to perform diagnosis. 
% The other subfigures also illustrate this situation, which fully demonstrates that our embedding representation unit can learn the correct histological feature representation for bags from a large amount of data. 

\begin{figure}[tb]
\centering\includegraphics[width=0.5\textwidth]{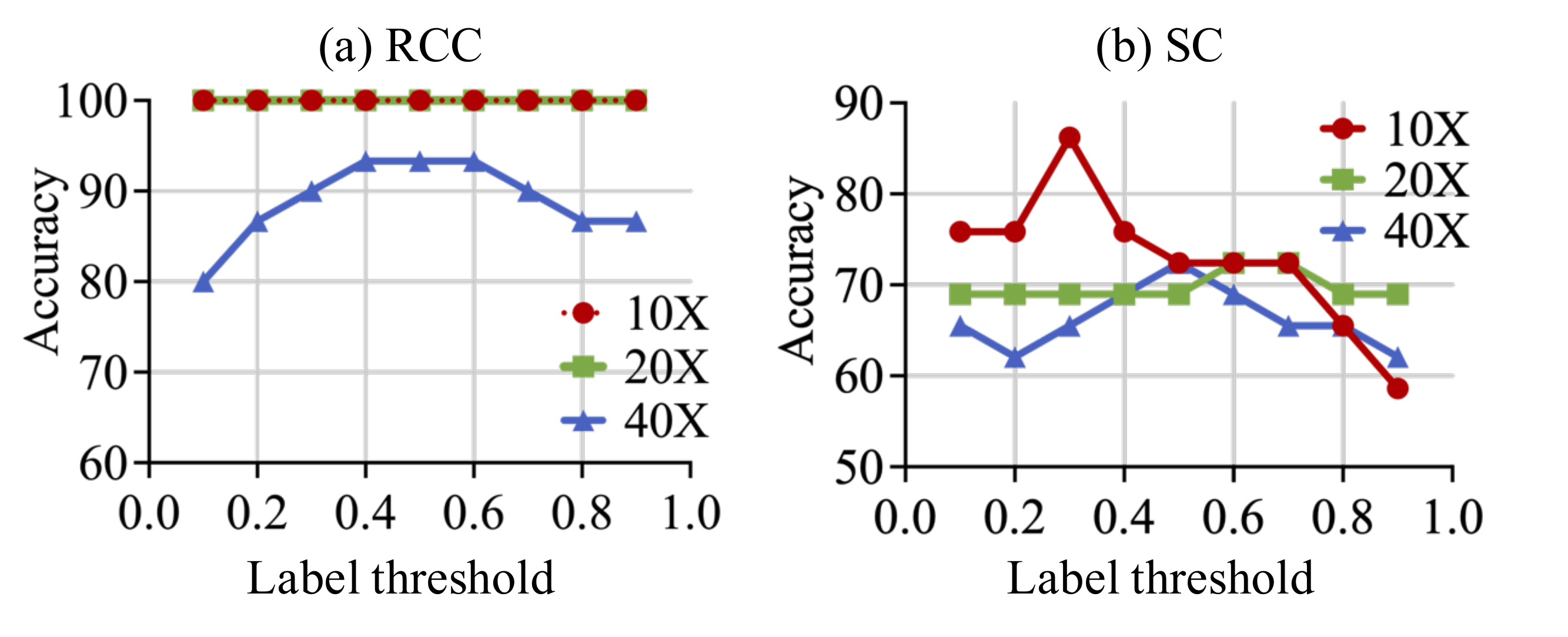}
\caption{The accuracy curve of D\&K using different label thresholds. The curves of $s = 1$ and $s = 2$ overlap in (a).}
\label{thrd}
\end{figure}

Next, we verify the performance of the bagging ensemble unit to convert bag-level histological features to slide-level features by the label threshold. 
In Figure \ref{thrd} (a), the number of histological features is relatively balanced in the RCC dataset, so the label threshold has little effect on the classification performance of the model and the initial value of 0.5 gives the best performance.
Due to the complexity of actual clinical data, the accuracy of the model on the SC dataset fluctuates in Figure \ref{thrd} (b). 
The values of the label threshold that achieve the best performance on the three scales are 0.3, 0.6, and 0.5. 
The reason is that the number of histological features in SC is imbalanced, and the label threshold will be lower with the increased number of constraint conditions. 
Nevertheless, the value of the label threshold can still be maintained at approximately 0.5, which also shows the robustness of our model. 

Therefore, a reasonable label threshold can convert continuous values to discrete values for a bag and obtain slide-level representations by a voting mechanism in the bagging ensemble unit. 
Through these experiments, we can conclude that the data-driven module can improve the representation ability of D\&K. Explicit histological features can help the model perform better in cancer subtype diagnosis. 

\begin{table*}[tb]
\centering
\caption{Comparison of the effectiveness of the knowledge-driven module and neural networks for multi-scale data on RCC and SC datasets.}
\footnotesize
\setlength{\tabcolsep}{1.5mm}{
%\resizebox{\linewidth}{!}{
\begin{tabular}{cc|ccccc|ccccc}
\toprule
 & & \multicolumn{5}{c}{RCC} & \multicolumn{5}{c}{SC}\\
\midrule
Refer to & Strategy & ACC & P & R & S & F1 & ACC & P & R & S & F1\\
\midrule
\midrule
\tabincell{c}{ConvNet \cite{COMPSAC2019}\\Tong et al. (2019)} & Average & 0.9917 & 0.9917 & 0.9917 & 0.9959 & 0.9917 & 0.6248 & 0.5937 & 0.5889 & 0.8053 & 0.5904\\
\tabincell{c}{AWMF-CNN \cite{cvpr2019}\\Tokunaga et al. (2019)} & Weight & 0.9950 & 0.9950 & 0.9950 & 0.9975 & 0.9950 & 0.6469 & 0.5441 & 0.5564 & 0.8097 & 0.5274\\
\tabincell{c}{MS-DA-MIL \cite{ms-da-mil}\\Hashimoto et al. (2020)} & Concatenation & 0.9642 & 0.9674 & 0.9642 & 0.9821 & 0.9641 & 0.5917 & 0.4029 & 0.5024 & 0.7781 & 0.4408\\
\midrule
\midrule
D\&K (Ours) & Knowledge & \textbf{1.0000} & \textbf{1.0000} & \textbf{1.0000} & \textbf{1.0000} & \textbf{1.0000} & \textbf{0.8966} & \textbf{0.9184} & \textbf{0.8889} & \textbf{0.9434} & \textbf{0.8984}\\
\bottomrule
\end{tabular}
}
\label{table2}
\end{table*}

\begin{figure*}[tb]
\centering\includegraphics[width=1\textwidth]{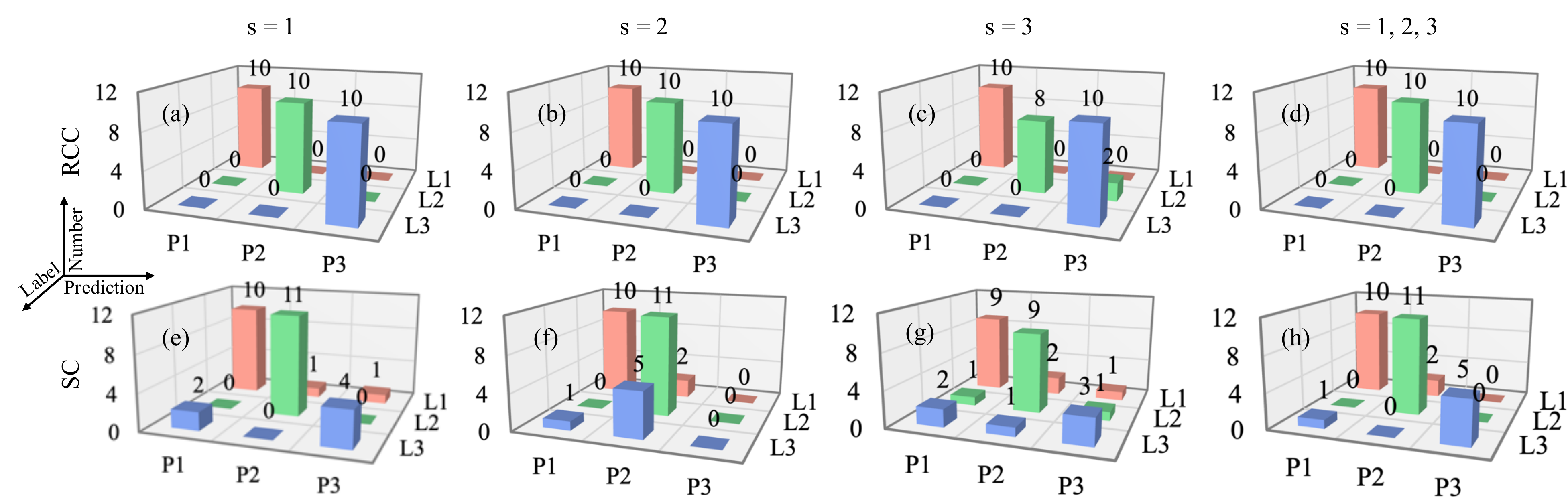}%0.48
\caption{Classification results of D\&K with the Gestalt principle under different scales. (a)-(d) and (e)-(h) represent the diagnosis results on $s = 1$, $s = 2$, $s = 3$, and $s = 1, 2, 3$ for RCC and SC datasets, respectively.} 
\label{heatmap}
\end{figure*}
\subsection{Benefit of Knowledge-Driven Module (Q3)}
This section verifies whether the knowledge-driven module based on Gestalt principles is effective for subtype classification and better than just using neural networks. 

First, we cannot directly remove the knowledge-driven module to obtain the subtype classification results due to the output of the data-driven module being histological features. Therefore, we can use the high-dimensional features extracted by the embedding representation unit for subtype classification. This design is the same as MIL using VGG16 in single-scale mode, so we can obtain this result from the literature \cite{mil-baseline}. 
It can be seen from the ACC indicators in Table \ref{table1} that the fusion of expert knowledge at a single scale can effectively improve classification accuracy, especially on the SC dataset.
Next, we extract histological features from single-scale images and fuse expert knowledge to illustrate the effectiveness of incorporating knowledge-driven modules.
Figure \ref{heatmap} shows a 3D confusion matrix of the subtype classification results, and the knowledge-driven module using the proximity and similarity principles performs well on the three single scales from Figure \ref{heatmap} (a)-(c) and (e)-(g). 
However, some predictions are incorrect on a single scale because the model does not use any histological features from multiple scales to analyze the contextual information. 
In Figure \ref{heatmap} (d) and (h), after combining the information from three scales by the Gestalt principle, it has a pathologist-like global to local diagnostic ability. 
Moreover, the subtype classification performance of the model on two datasets is further improved, especially for BD in SC. 
This shows that the model in this paper can consider multi-scale information to provide scientific diagnosis suggestions using the Gestalt principle, which is an effective way to integrate the knowledge-driven module. 

Second, we consult three popular methods for multi-scale fusion using neural networks. 
ConvNet uses sum and means to obtain a unified representation from multi-scale features. 
AWMF-CNN uses a weight mechanism to combine important scale features, and MS-DA-MIL combines multi-scale features by a concatenation function. 
For the sake of fairness, the embedding for the compare algorithms at each scale is obtained by the data-driven part of D\&K (before sigmoid), and the results are predicted by the classification layer consisting of fully connected neural networks with attention.
As shown in Table \ref{table2}, the multi-scale models using neural networks cannot combine the contextual information very well, and the classification performance is average. 

Overall, the experimental results show that our model obtains more significant performance than baselines by adding expert knowledge. 
This indicates that the knowledge-driven module consisting of proximity and similarity principles of Gestalt has an intelligible inference process for subtype classification. 

\begin{figure*}[t]
\centering\includegraphics[width=1\textwidth]{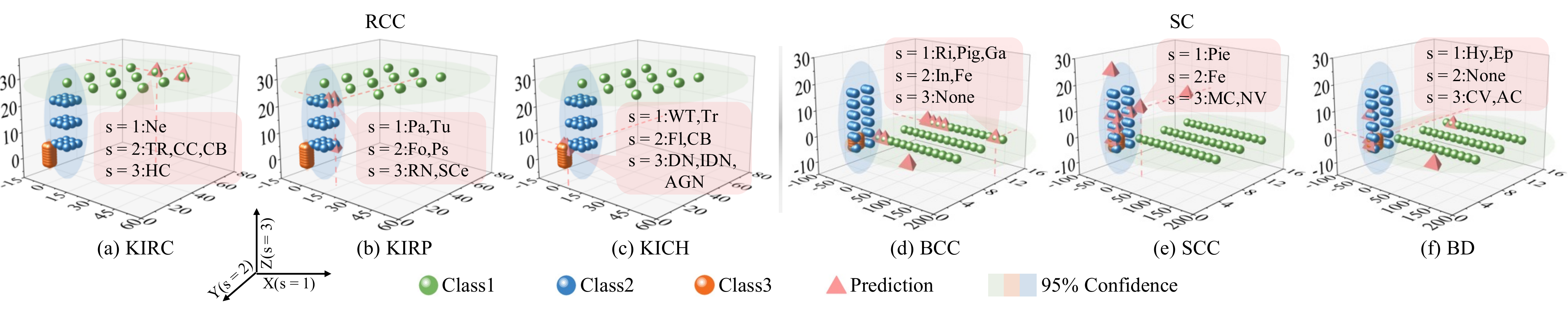}%0.48
\caption{Mapping visualization of histological features in the 3D expert knowledge space. (a)-(c) and (d)-(f) represent the histological features on different scales and final classification results for RCC and SC datasets.} 
\label{3d}
\end{figure*}

\begin{figure*}[tb]
\centering\includegraphics[width=1\textwidth]{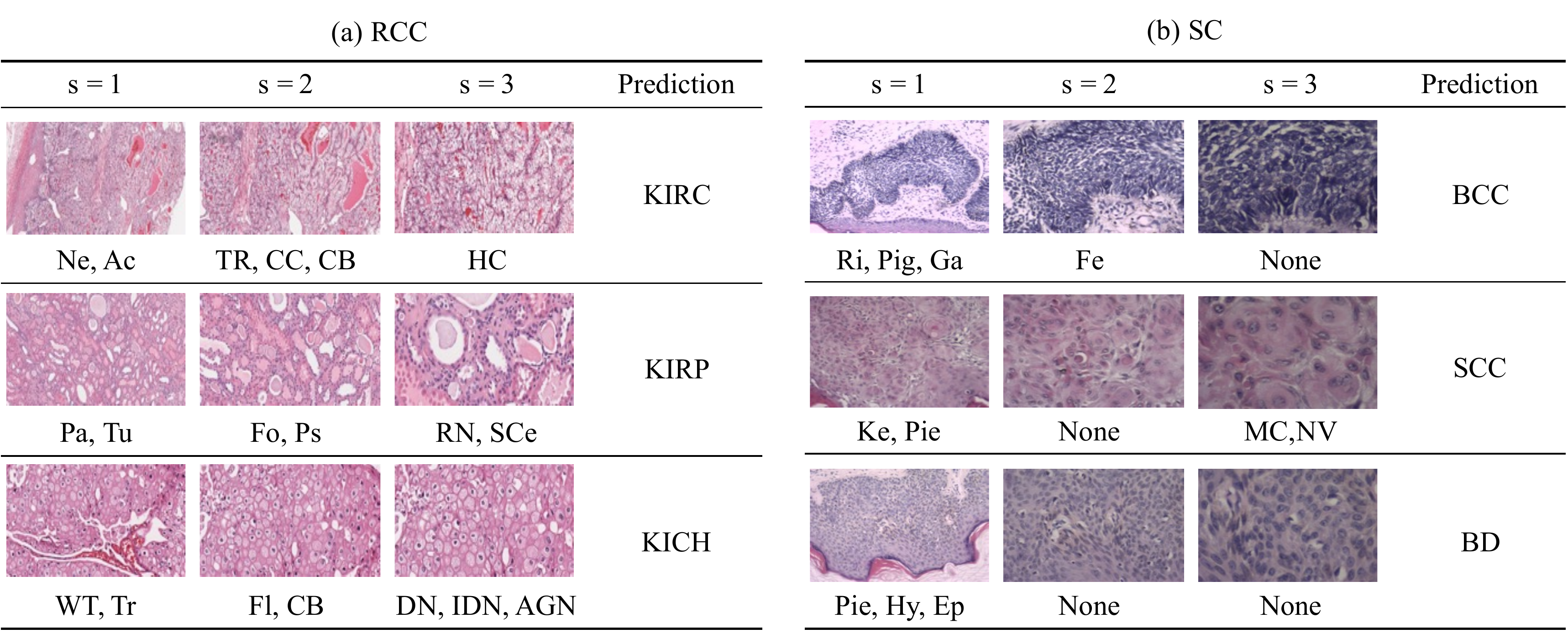}
\caption{Diagnosis results of several samples.} 
\label{result}
\end{figure*}

\subsection{Visualization of Diagnosis (Q4)}
For an in-depth analysis of D\&K, we visualize the prediction and expert knowledge distribution with triangles and balls in 3D space. We want to use qualitative experiments to explore how the model obtains high classification performance and credible results by explicit representation and intelligible inference. 

For explicit representation, the samples are converted from images to histological features through a data-driven module and then mapped to different locations in 3D expert knowledge space. 
In Figure \ref{3d}, the samples shown by the triangles reveal the histological features of the cancer subtypes at each scale, and the model maps them to different locations in the 3D space based on these features. 
The histological features generated by the data-driven module intuitively visualize and illustrate what the model is based on to give diagnostic conclusions, which can help doctors make better decisions. Therefore, our model has an explicit representation ability for improving performance on cancer subtype diagnosis.

For intelligible inference, the core idea is to leverage the knowledge-driven module by the Gestalt principle. 
The expert knowledge of each subtype is modeled into a very close position by the proximity principle, and the predictions obtained by the data-driven module are mapped into the corresponding region for calculating distance. 
The distance calculated by the similarity principle can reflect the correlations between prediction points and expert knowledge, and the model can give the final judgment results of the subtypes based on it. 
As shown in Figure \ref{3d}, most of the predicted points are correctly mapped into the regions where they belong to the subtypes (the regions in the graph are formed with 95\% confidence). 
Moreover, there are some overlapping points, making it difficult to observe. 
It is worth noting that some results near SCC are also predicted as BD because they can be judged as BD when the histological features of \emph{Ep} appear on the scale of $s=1$. 

In addition, we visualize the outputs of our model to show credible diagnostic results. There are six samples under three scales, including histological features and subtype prediction. First, D\&K extracts histopathological features from the slides of three scales for each sample by the data-driven module. Then, D\&K measures the similarity between the histological prediction and expert knowledge with the knowledge-driven module for classification. Figure \ref{result} shows that the model identified the histological features, and accordingly, it was able to distinguish the correct subtypes by using expert knowledge. In general, D\&K gives the histological lesion information at each scale, and it is a high-performance and intelligible model for the doctor.

In conclusion, the visualization process shows details of D\&K overlaid with how to combine the data-driven and knowledge-driven modules to obtain a high-performance and credible result of subtype diagnosis.

\section{Conclusion}
In this paper, we propose a novel D\&K model that reconstructs the diagnosis process of cancer subtypes using two core modules. 
In the data-driven module, embedding representation and bagging ensemble units are used to analyze histological features at different scales on the slide. 
After that, the knowledge-driven module builds the 3D expert knowledge space with the proximity principle, and histological features are mapped into this space with the similarity principle to diagnose subtypes. 
We illustrate and prove the effectiveness and intelligibility of D\&K on in-house and real-world datasets from four aspects. 
It outperforms the state-of-the-art methods on histopathological slide subtype classification. 
For diagnostic results, the model can give histological features at each scale and diagnose cancer subtypes based on changes in these features. 
Compared with other algorithms, D\&K gains a high classification accuracy and explores the possibility of credible results. 
Since this paper is a pioneering work, the histological features need to be labeled with more content. However, the model will become more accessible and easier to use as collaboration continues. 
For future work, we will focus on refining and extending D\&K to other cancer subtype classification. 

\section*{Acknowledgments}
This work is partially supported in part by the National Natural Science Foundation of China under grants Nos. 61902145, 61976102, and U19A2065; the International Cooperation Project under grant No. 20220402009GH; the National Key R\&D Program of China under grants Nos. 2021ZD0112501 and 2021ZD0112502; and the Innovation Project of New Generation Information Technology under grant No. 2021ITA05001; the China Scholarship Council Project (202206170090).

%Bibliography
\bibliographystyle{unsrt}  
\bibliography{references}  

\clearpage

\appendix

\section{Expert Knowledge for D\&K}
%\vskip3pt
In this section, we introduce expert knowledge of RCC and SC in detail, and we describe the histopathological features of each subtype at each scale. 

First, we introduced the diagnosis knowledge of RCC at the 's = 1' scale in Section 4.3. In Table \ref{table-rc2}, there are six histological features on the scale of 's = 2': Thin-Reticulate (TR) and Clear-Cell (CC) are the features of KIRC, Foamy (Fo) and Psammoma (Ps) are the features of KIRP; and Flocculence
(Fl) and Clear-Boundary (CB) are the features of KICH. Table \ref{table-rc3} shows the histological features on the scale of 's = 3': Homogenous-Chromatin (HC) is the histological feature of KIRC; Round-Nucleus (RN) and Small-Cell (SCe) are the histological features of KIRP; and Double-Nucleus (DN), Irregular-Dark-Nucleus (IDN), and Air-Gap-Nucleus (AGN) are the histological features of KICH.

Second, as shown in Table \ref{table-sc1}, they are the diagnosis knowledge for SC. It has nine histological features related to skin cancer subtypes on a scale of 's = 1'. There are five histological features of BCC: Stripe (St), Ribbon (Ri), Cribriform (Cr), Pigment (Pig), and Gap (Ga); two histological features of SCC: Keratinization (Ke) and Piece (Pie); and two histological features of BD: Hypertrophy (Hy) and Epidermis (Ep). In Table \ref{table-sc2}, there are four histological features on the scale of 's = 2': Interstitial (In) and Fence (Fe) are the features of BCC, and Implicate-Vessel (IV) and Implicate-Adnexa (IA) are the features of SCC and BD, respectively. Table \ref{table-sc3} shows the histological features on the scale of 's = 3': Intercellular-Bridge (IB), More-Cytoplasm (MC), and Nuclear-Vacuolation (NV) are the histological features of SCC; Cytoplasm-Vacuolation (CV), and Alien-Cell (AC) are the features of BD. BCC has no features on this scale. It should be noted that all of the above combinations represent the knowledge required by experts in making a diagnosis. However, in the data-driven part, a slice can only be predicted by the model in one possible combination.

In general, we combine these features into various possible forms, which conform to expert knowledge for diagnosing RCC and SC subtypes. At the same time, our histopathological images are also annotated and trained according to these combinations in the training phase.

\section{Time and Space Complexity of D\&K}

Algorithm \ref{train_algorithm} and Algorithm \ref{prediction_algorithm} are pseudocodes of the D\&K model. We first analyze the time complexity and then analyze the space complexity as follows:

The time complexity should be `$3*E*N$' according to the pseudocode definition in the data-driven module. However, $E$ represents the training items, and it is a constant. In addition, $N$ or $N^{'}$ means the sample size, which is an indeterminate variable. Therefore, the time complexity of this module is $O(n)$ according to the calculation rules. Therefore, as in the knowledge-driven module, the time complexity of this module is $O(n)$.

For space complexity, they are both $O(n)$ in these two modules since D\&K will store the input samples $N$ or $N^{'}$ as a resource when it starts running.

\begin{table}[htbp]
\centering
\caption{The expert knowledge of SC on $s = 1$.}
\setlength{\tabcolsep}{1.8mm}{
\begin{tabular}{cccccccccc}
\toprule
\rotatebox{0}{Subtypes} & \rotatebox{0}{St} & \rotatebox{0}{Ri} & \rotatebox{0}{Cr} & \rotatebox{0}{Pig} & \rotatebox{0}{Ga} & \rotatebox{0}{Ke} & \rotatebox{0}{Pie} & \rotatebox{0}{Hy} & \rotatebox{0}{Ep} \\
\midrule
\midrule
\multirow{31}*{\rotatebox{0}{BCC}} & 1& 1& 1& 1& 1& 0& 0& 0& 0 \\
		 & 1& 1& 1& 1& 0& 0& 0& 0& 0 \\
		 & 1& 1& 1& 0& 1& 0& 0& 0& 0 \\
		 & 1& 1& 1& 0& 0& 0& 0& 0& 0 \\
		 & 1& 1& 0& 1& 1& 0& 0& 0& 0 \\
		 & 1& 1& 0& 1& 0& 0& 0& 0& 0 \\
		 & 1& 1& 0& 0& 1& 0& 0& 0& 0 \\
		 & 1& 1& 0& 0& 0& 0& 0& 0& 0 \\
		 & 1& 0& 1& 1& 1& 0& 0& 0& 0 \\
		 & 1& 0& 1& 1& 0& 0& 0& 0& 0 \\
		 & 1& 0& 1& 0& 1& 0& 0& 0& 0 \\
		 & 1& 0& 1& 0& 0& 0& 0& 0& 0 \\
		 & 1& 0& 0& 1& 1& 0& 0& 0& 0 \\
		 & 1& 0& 0& 1& 0& 0& 0& 0& 0 \\
		 & 1& 0& 0& 0& 1& 0& 0& 0& 0 \\
		 & 1& 0& 0& 0& 0& 0& 0& 0& 0 \\
		 & 0& 1& 1& 1& 1& 0& 0& 0& 0 \\
		 & 0& 1& 1& 1& 0& 0& 0& 0& 0 \\
		 & 0& 1& 1& 0& 1& 0& 0& 0& 0 \\
		 & 0& 1& 1& 0& 0& 0& 0& 0& 0 \\
		 & 0& 1& 0& 1& 1& 0& 0& 0& 0 \\
		 & 0& 1& 0& 1& 0& 0& 0& 0& 0 \\
		 & 0& 1& 0& 0& 1& 0& 0& 0& 0 \\
		 & 0& 1& 0& 0& 0& 0& 0& 0& 0 \\
		 & 0& 0& 1& 1& 1& 0& 0& 0& 0 \\
		 & 0& 0& 1& 1& 0& 0& 0& 0& 0 \\
		 & 0& 0& 1& 0& 1& 0& 0& 0& 0 \\
		 & 0& 0& 1& 0& 0& 0& 0& 0& 0 \\
		 & 0& 0& 0& 1& 1& 0& 0& 0& 0 \\
		 & 0& 0& 0& 1& 0& 0& 0& 0& 0 \\
		 & 0& 0& 0& 0& 1& 0& 0& 0& 0 \\
\midrule
\multirow{3}*{\rotatebox{0}{SCC}} & 0& 0& 0& 0& 0& 1& 1& 0& 0 \\
		 & 0& 0& 0& 0& 0& 1& 0& 0& 0 \\
		 & 0& 0& 0& 0& 0& 0& 1& 0& 0 \\
\midrule
\multirow{3}*{\rotatebox{0}{BD}} & 0& 0& 0& 0& 0& 0& 0& 1& 1 \\
		 & 0& 0& 0& 0& 0& 0& 0& 1& 0 \\
		 & 0& 0& 0& 0& 0& 0& 0& 0& 1 \\
\bottomrule
\end{tabular}
}
\label{table-sc1}
\end{table}

\begin{table}[htbp]
\centering
\footnotesize
\caption{The expert knowledge of RCC on $s = 2$.}
\setlength{\tabcolsep}{3.5mm}{
\begin{tabular}{ccccccc}
\toprule
\rotatebox{0}{Subtypes} & \rotatebox{0}{TR} & \rotatebox{0}{CC} & \rotatebox{0}{Fo} & \rotatebox{0}{Ps} & \rotatebox{0}{Fl}& \rotatebox{0}{CB}\\
\midrule
\midrule
\multirow{3}*{\rotatebox{0}{KIRC}} & 1& 1& 0& 0& 0& 0 \\
		 & 1& 0& 0& 0& 0& 0 \\
		 & 0& 1& 0& 0& 0& 0 \\
\midrule
\multirow{3}*{\rotatebox{0}{KIRP}} & 0& 0& 1& 1& 0& 0 \\
		 & 0& 0& 1& 0& 0& 0 \\
		 & 0& 0& 0& 1& 0& 0 \\
\midrule
\multirow{3}*{\rotatebox{0}{KICH}} & 0& 0& 0& 0& 1& 1 \\
		 & 0& 0& 0& 0& 1& 0 \\
		 & 0& 0& 0& 0& 0& 1 \\
\bottomrule
\end{tabular}
}
\label{table-rc2}
\end{table}

\begin{table}[htbp]
\centering
\footnotesize
\caption{The expert knowledge of RCC on $s = 3$.}
\setlength{\tabcolsep}{3mm}{
\begin{tabular}{ccccccc}
\toprule
\rotatebox{0}{Subtypes} & \rotatebox{0}{HC} & \rotatebox{0}{RN} & \rotatebox{0}{SCe} & \rotatebox{0}{DN} & \rotatebox{0}{IDN}& \rotatebox{0}{AGN}\\
\midrule
\midrule
\multirow{1}*{\rotatebox{0}{KIRC}} & 1& 0& 0& 0& 0& 0 \\
\midrule
\multirow{3}*{\rotatebox{0}{KIRP}} & 0& 1& 1& 0& 0& 0 \\
		 & 0& 1& 0& 0& 0& 0 \\
		 & 0& 0& 1& 0& 0& 0 \\
\midrule
\multirow{6}*{\rotatebox{0}{KICH}} & 0& 0& 0& 1& 1& 1 \\
		 & 0& 0& 0& 1& 1& 0 \\
		 & 0& 0& 0& 1& 0& 1 \\
		 & 0& 0& 0& 0& 1& 1 \\
		 & 0& 0& 0& 0& 1& 0 \\
		 & 0& 0& 0& 0& 0& 1 \\
\bottomrule
\end{tabular}
}
\label{table-rc3}
\end{table}

\begin{table}[htbp]
\centering
\caption{The expert knowledge of SC on $s = 2$.}
\setlength{\tabcolsep}{5mm}{
\begin{tabular}{ccccc}
\toprule
\rotatebox{0}{Subtypes} & \rotatebox{0}{In} & \rotatebox{0}{Fe} & \rotatebox{0}{IV} & \rotatebox{0}{IA} \\
\midrule
\midrule
\multirow{3}*{\rotatebox{0}{BCC}} & 1& 1& 0& 0 \\
		 & 1& 0& 0& 0 \\
		 & 0& 1& 0& 0 \\
\midrule
\multirow{1}*{\rotatebox{0}{SCC}} & 0& 0& 1& 0 \\
\midrule
\multirow{1}*{\rotatebox{0}{BD}} & 0& 0& 0& 1 \\
\bottomrule
\end{tabular}
}
\label{table-sc2}
\end{table}

\begin{table}[htbp]
\centering
\caption{The expert knowledge of SC on $s = 3$.}
\setlength{\tabcolsep}{3.5mm}{
\begin{tabular}{cccccc}
\toprule
\rotatebox{0}{Subtypes} & \rotatebox{0}{IB} & \rotatebox{0}{MC} & \rotatebox{0}{NV} & \rotatebox{0}{CV} & \rotatebox{0}{AC}\\
\midrule
\midrule
\multirow{7}*{\rotatebox{0}{SCC}} & 1& 1& 1& 0& 0 \\
		 & 1& 1& 0& 0& 0 \\
		 & 1& 0& 1& 0& 0 \\
		 & 1& 0& 0& 0& 0 \\
		 & 0& 1& 1& 0& 0 \\
		 & 0& 1& 0& 0& 0 \\
		 & 0& 0& 1& 0& 0 \\
\midrule
\multirow{3}*{\rotatebox{0}{BD}} & 0& 0& 0& 1& 1 \\
		 & 0& 0& 0& 1& 0 \\
		 & 0& 0& 0& 0& 1 \\
\bottomrule
\end{tabular}
}
\label{table-sc3}
\end{table}

\end{document}